\documentclass[letterpaper,10pt,conference]{ieeeconf}
\usepackage{algpseudocode}
\usepackage{graphics} 
\usepackage{epsfig} 
\usepackage{mathptmx} 
\usepackage{times} 
\usepackage{amsmath} 
\usepackage{amssymb}  
\usepackage[ruled,vlined]{algorithm2e}
\usepackage{subfig}
\usepackage{hyperref,color}
\usepackage{graphicx} 
\usepackage{soul}

\title{\large \bf A Model-free Deep Reinforcement Learning Approach To Maneuver A Quadrotor Despite Single Rotor Failure }
\author{Paras Sharma, Prithvi Poddar, and P.B. Sujit%
\thanks{Paras Sharma is with IIIT Delhi, New Delhi -- 110020, India. email: paras17250@iiitd.ac.in}
\thanks{Prithvi Poddar and P.B. Sujit are with the Department of Electrical Engineering and Computer Science at IISER Bhopal, Bhopal -- 462066, India. email: (prithvid17,sujit)@iiserb.ac.in}
}

\IEEEoverridecommandlockouts                              
                                                          
\begin{document}
\maketitle
\begin{abstract}
Ability to recover from faults and continue mission is desirable for many quadrotor applications. The quadrotor's rotor may fail while performing a mission and it is essential to develop recovery strategies so that the vehicle is not damaged. In this paper, we develop a model-free deep reinforcement learning approach for a quadrotor to recover from a single rotor failure. The approach is based on  Soft-actor-critic that enables the vehicle to hover, land, and perform complex maneuvers. Simulation results are presented to validate the proposed approach using a custom simulator. The results show that the proposed approach achieves hover, landing, and  path following in 2D and 3D. We also show that the proposed approach is robust to wind disturbances. 
\end{abstract}

\section{Introduction} \label{sec:intro}

Quadrotors  are used in various applications ranging across military purposes~\cite{article}, automating deliveries~\cite{KELLERMANN2020100088}, surveillance~\cite{7759885}, search and rescue~\cite{POLKA2017748}, etc.  Quadrotors  use the differential thrust created by the rotors to fly. The rotors are the most essential part of a quadrotor and the failure of even a single rotor can  lead to a catastrophic crash. Hence, there is an impending need for developing strategies to recover a quadrotor with a rotor failure and possibly enable it to retain its maneuvering capability so that the mission can still be continued especially in persistent monitoring applications. 


There have been several works that focus on developing recovery strategies under single rotor failures (SRF). There are two possible solution under SRF -- (i) modify the configuration of the quadrotor and (ii) develop efficient fault tolerant controllers. The first approach in used in  \cite{DSCC2018-9197, DSCC2016-9897, robotics7040065, STABILITYANDCONTROL} by using tilting rotor configuration. Under SRF, the remaining three rotors change their orientation by rotating along the axis passing through the arm of the UAV. 
 In \cite{avant2018dynamics} a morphing quadrotor is designed that changes its configuration by changing the arm configuration. \cite{8968099} presents a detailed aerodynamic model of the quadrotor and the propellers and provides a fault tolerant strategy using tilting rotors and retractable arms. However, adding additional servos to the quadrotor further increases the complexity of the model and also possibility of additional failures.
 
 Several nonlinear controllers have also been developed to recover from SRF. In \cite{9197239}, a cascade controller is designed to recover the flight from any arbitary position with 3 rotors. In \cite{lanzon2014flight}, a $H_\infty$ controller is designed to recover from rotor failures. In \cite{lippiello2014emergency}, a backstepping approach to handle single rotor failure by stopping the rotor opposite to the failed rotor, essentially forming a birotor configuration and performing an emergency landing. \cite{lee2020fail} uses a configurable centre of mass system for stabilization to achieve similar performance of that of a 4 rotor system.  A vision based approach was in \cite{sun2021autonomous}, where the features are obtained using a event driven camera and state-estimation along with a fault-tolerant controller are desiged for recovery. In  \cite{ hou2020nonsingular},  a non singular sliding mode controller is designed to handle SRF in quadrotors. In \cite{de2015unified}, an iterative optimal control algorithm is utilized to design the policies for the vehicle to stabilize under single and double rotor failures.

An alternative approach to recover from SRF is to reinforcement learning (RL) algorithms. In \cite{ETH_Original_paper}, the authors have shown how reinforcement learning controls a quadrotor even with very high level external disturbances. In \cite{fei2020learn}, RL is used to a fault tolerant controller that recovers the vehicle in hover mode and its performance is better than the traditional fault tolerant methods. In \cite{arasanipalai2020mid} RL and convolutional neural networks are used to achieve recoverable hover condition after rotor failures.  

In this paper, we advance the approaches developed in \cite{fei2020learn,arasanipalai2020mid} to not only achieve hover, but also achieve maneuverability. We  use Soft Actor-Critic (SAC) deep reinforcement learning algorithm to achieve fault tolerance. Further, we use a model-free  methodology compared to model-based \cite{fei2020learn,arasanipalai2020mid} which allows the system to learn any kind of uncertainties and unmodelled errors which can be an issue in model-based learning. The main contributions of this paper are
\begin{itemize}[leftmargin=*]
    \item Formulate the problem as a deep reinforcement learning problem through the use of SAC 
    \item Develop a simulator for SRF
    \item Show that SAC-based RL framework prevents a quadrotor from crashing and  provides maneuverability along all the 3 axes, with the remaining three active rotors.
    \item Show that the quadrotor with a SRF can preform three types of functions \textit{(a)} hover, \textit{(b)} land and \textit{(c)} path following in 3D.   
\end{itemize}


\section{Problem Formulation} 
\label{sec:problem}
\begin{figure}
    \centering
    \includegraphics[width=7cm]{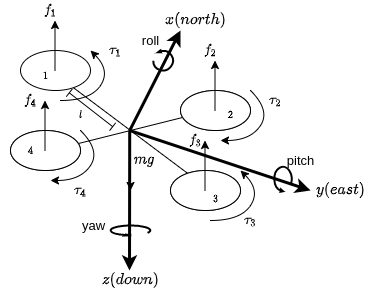}
    \caption{The body frame of a quadrotor. We consider a NED (North East Down) frame of reference. The rotors 1, 2, 3 and 4, at the ends of each arm of length $l$, generate upward forces $f_1, f_2, f_3$ and $f_4$, while rotating at angular velocities of $\omega_1. \omega_2, \omega_3$ and $\omega_4$. Rotation along the north axis is \textit{roll}, rotation along the east axis is \textit{pitch} and rotation along the down axis is \textit{yaw}.}
    \label{fig:drone_schematic}
\end{figure}
Quadrotors use differential thrust generated from fixed rotors for stabilization and control. Four control inputs are required for attitude control -- roll, pitch and yaw and thrust for altitude control. Due to presence of rotors, they are susceptible to rotor failure. In Fig. \ref{fig:drone_schematic} we see that the rotors generate thrust ($f_1, f_2, f_3, f_4$) in the opposite direction of the weight of the quadrotor $mg$. The thrust from each rotor generates a torque $f_i\times l$ on the centre of mass of the quadrotor. When the thrusts on each rotor are equal, the resultant torque is 0, enabling the vehicle move only along the $z-axis$. The thrusts ($f_1, f_2, f_3, f_4$) on each rotor, can be controlled individually to generate thrust differentials, which in turn control the roll and the pitch angles. The angular momentum of the quadrotor is mitigated by have a configuration of two clockwise rotating rotors and two anti-clockwise rotating rotors. 
The diagonally opposite rotors spin in the same direction and yaw control is achieved by increasing or decreasing the angular velocities of the diagonally opposite rotors~\cite{DynamicsModelingControl}.

In this paper, we study the case of a single rotor failure (SRF) where we assume that one of the rotors has completely stopped spinning and is no longer capable of producing thrust. Such a situation causes imbalance in the net angular momentum and torque acting on the quadrotor. Although, the roll, pitch and thrust controls can be retained by accurate modeling of the system, but yaw control cannot regained. Failure of a rotor also leads to the change in the dynamics of the quadrotor and this change in itself depends on the physical model of the drone (affected by parameters like the length of the arm, moment of inertia, weight of the drone, etc). Designing an adaptive control system becomes challenging in these cases. The controllers designed in \cite{9197239}-\cite{hou2020nonsingular} can be used for this purpose, but they are not robust to any parameter variations in the model. Therefore, in this paper, we design the controller using a model-free approach through deep reinforcement learning  with relative ease. This controller recovers the quadrotor from a SRF and retains its maneuverability capabilities even though yaw control still remains unavailable.

We assume that the quadrotor has a fault detection to detect the fault. The RL controller executes parallely to the vehicle low-level controller and takes over the low level controller one the fault is detected. We assume that SRF is detected using any of the approaches given in ~\cite{en12061139}.

\section{Methodology} \label{sec:methodology}

We use Soft-Actor-Critic algorithm \cite{SAC_main} to determine policies for the quadrotor under SRF. Soft Actor Critic (SAC) is an off-policy algorithm that has entropy regularization which helps in exploration-exploitation trade-off. The SAC\cite{SAC_main} performs better than other popular DRL algorithms such as DDPG \cite{DDPG}, PPO\cite{PPO}, SQL\cite{SQL} and TD3\cite{TD3}.

\subsection{Training Objective}
The objective is to make the quadrotor reach the goal position after the fault has occurred. The DRL controller needs to learn to go to a defined goal position. If the goal position is stationary, then the vehicle is hover. If the goal position is in motion, then the quadrotor should track it to minimize the error between the quadrotor position and the goal position.  We will now define the observation state and the reward function. 

\subsection{Observation State} \label{observation_state}
The observation state $s_t$ at any time $t$ has 22 state variables. They are
\begin{itemize}[leftmargin=*]
    \item Position error, $\{ \bar{x}_t, \bar{y}_t, \bar{z}_t\}$, where, $ \bar{x}_t  = x_{goal} - x_{t}, \bar{y}_t  = y_{goal} - y_{t},  \bar{z}_t  = z_{goal} - z_{t}$. 
    \item Full Rotation Matrix, $R_t = R_z(\mathbb{Y})R_y(\mathbb{P})R_x(\mathbb{R})$, where
    
    {\tiny
    $R_z(\mathbb{Y}) = \begin{bmatrix}
            \cos(\mathbb{Y}) & -\sin(\mathbb{Y}) & 0 \\
            \sin(\mathbb{Y}) & \cos(\mathbb{Y}) & 0 \\
            0 & 0 & 1
            \end{bmatrix}$
    $R_y(\mathbb{P}) = \begin{bmatrix}
            \cos(\mathbb{P}) & 0 & \sin(\mathbb{P}) \\
            0 & 1 & 0 \\
            -\sin(\mathbb{P}) & 0 & \cos(\mathbb{P})
            \end{bmatrix}$
            
    $R_x(\mathbb{R}) = \begin{bmatrix}
            1 & 0 & 0 \\
            0 & \cos(\mathbb{R}) & -\sin(\mathbb{R}) \\
            0 & \sin(\mathbb{R}) & \cos(\mathbb{R})\\
            \end{bmatrix}$
    }\\
    $\mathbb{Y}, \mathbb{P}, \mathbb{R}$ are the yaw, pitch and roll angles respectively.
    \item Linear velocity, \{$v^x_t, v^y_t, v^z_t$\}
    \item Angular velocity, \{$\omega^x_t, \omega^y_t,  \omega^z_t$\}
    \item RPM values of rotors at time $t$, \{$\tau_t^1, \tau_t^2, \tau_t^3, \tau_t^4$\}. Our model generates the change in PWM values rather than the absolute PWMs which motivates the need of current RPM values in the observation space.
\end{itemize}
In order to reduce complexity of the state space for faster on-board computation, we do not consider higher order derivatives of the motion. 

\subsection{Reward Function} \label{reward_function}
We have devised a simplistic reward function. As the quadrotor moves away from the goal, the vehicle received negative reward. The reward $r_t$ at time $t$ is given as 
\begin{eqnarray}
    r_t &=& r_1 + r_2,\\
    r_1 &=& -c_1\tanh(c_2\sqrt{\bar{x_t}^2 + \bar{y_t}^2 + \bar{z_t}^2})\\
    r_2 &=& -\sum_{i=1}^4\frac{|\tau_{t-1}^i - \tau_t^i|}{c_3}
\end{eqnarray}
where $c_1$, $c_2$ and $c_3$ are constants, $r_1$ is the responsible for maintaining the position of the quadrotor at the desired goal position. We use of the $\tanh$ function because it grows rapidly near the origin, which would provide greater incentive for the model to minimize the positional error. It also saturates asymptotically as the error increases, hence penalising all the high positional errors equally. $r_2$ moderates rapid changes in the angular velocities of the rotors. During initial experiments, it was found that without $r_2$, the model tends to change the angular velocities too rapidly. This would provide stability during the length of a training episode, but during testing for longer time periods, these rapid changes eventually lead the quadrotor to be unstable. In this paper, we use $c_1=10$, $c_2=0.2$ and $c_3=10$.

\begin{figure*}
    \centering
  \subfloat[]{\label{fig:critic_neteork} \includegraphics[width=5cm]{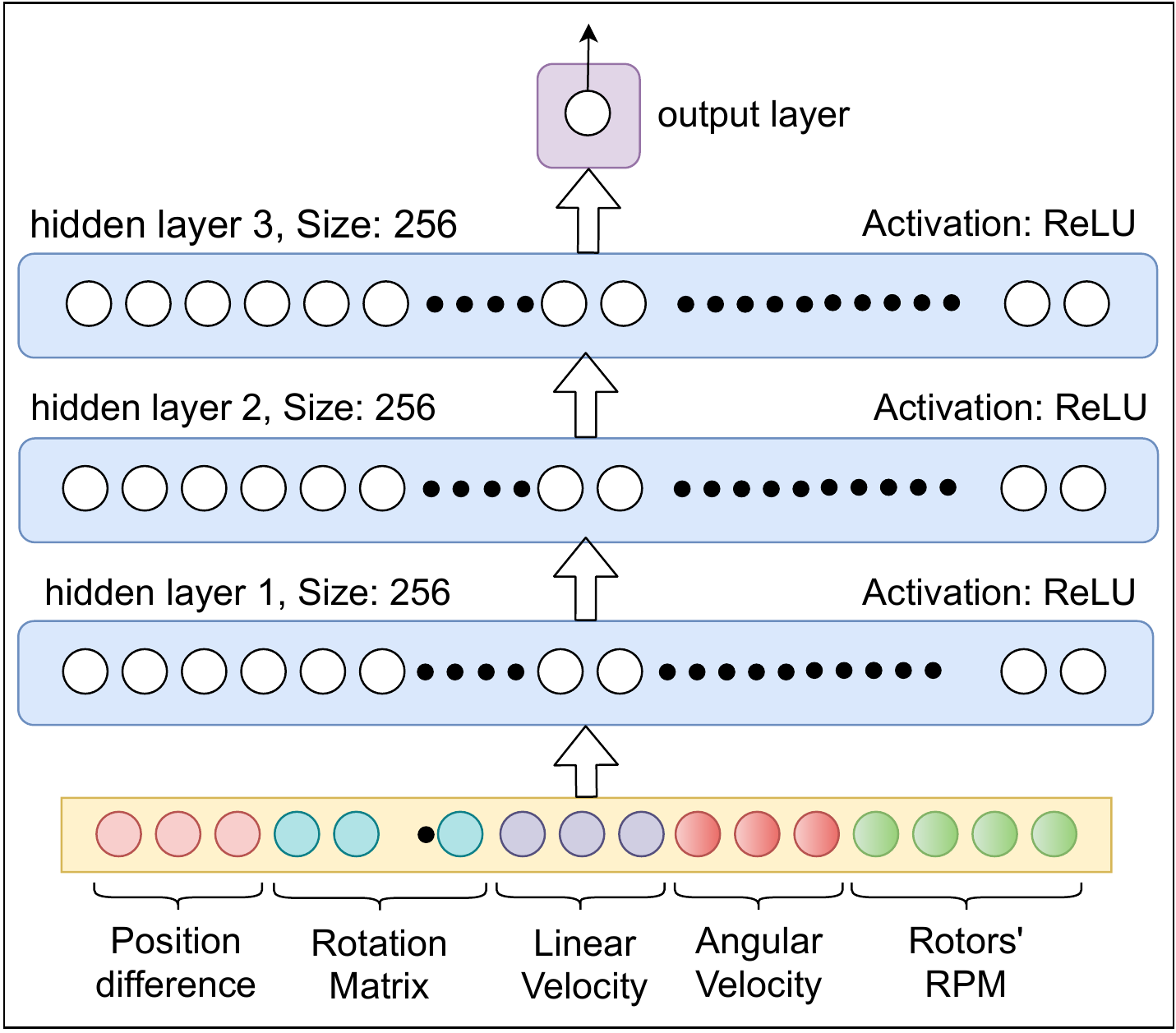}}\hspace{2mm}
  \subfloat[]{\label{fig:actor_network}\includegraphics[clip,trim = 0mm 0mm 5mm 0mm, width=6cm]{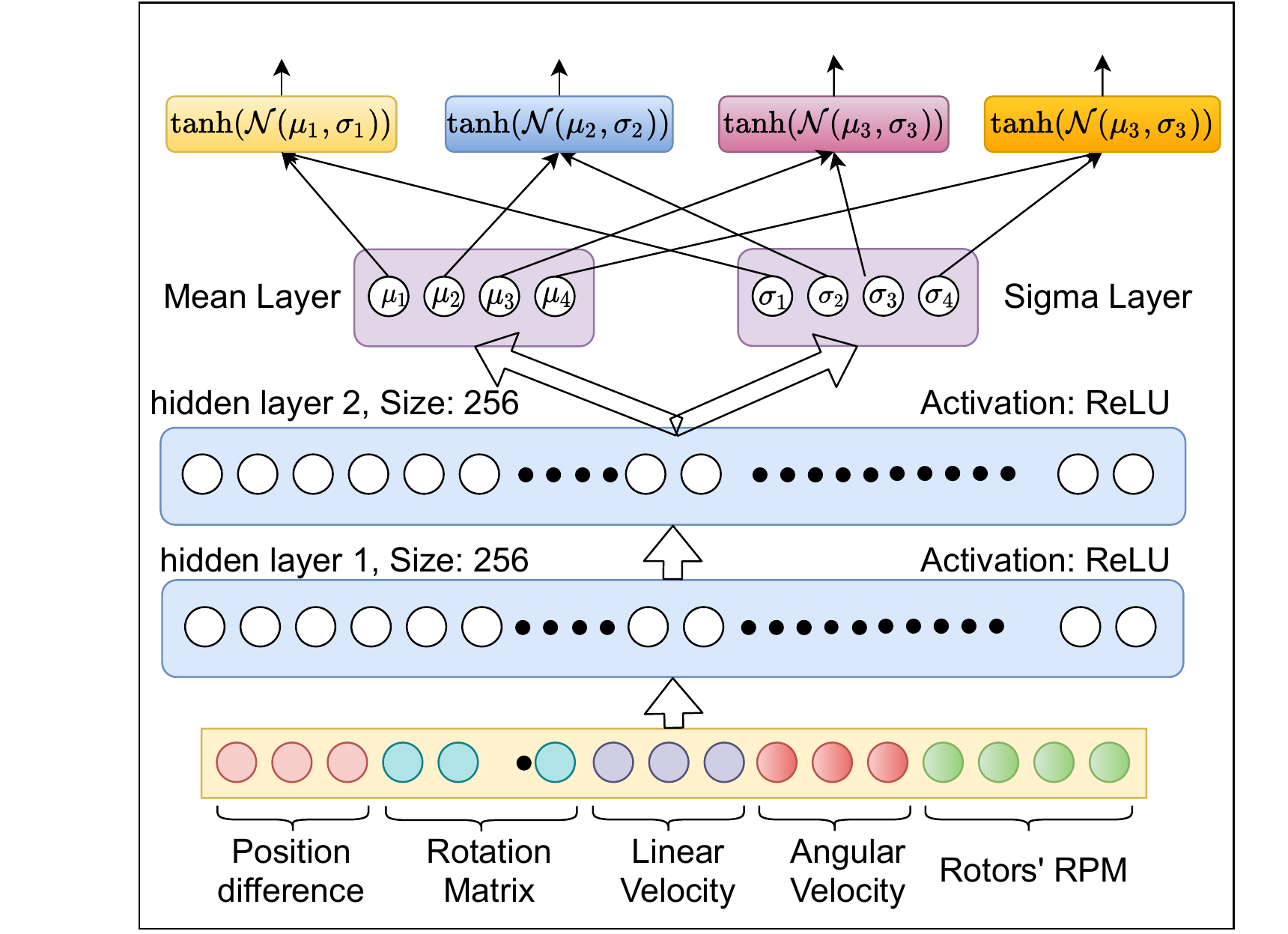}}
  \subfloat[]{\label{fig:hyperparam} \includegraphics[width=5cm]{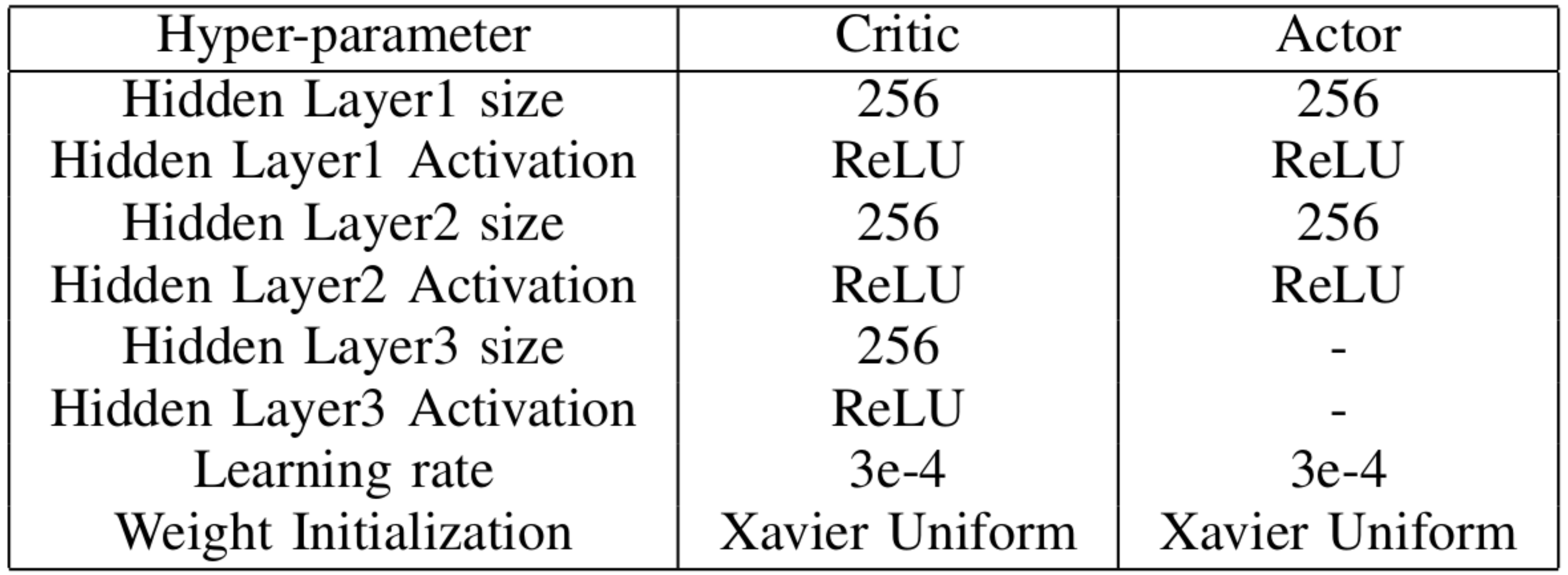}}
    \caption{(a) Critic Network: Three fully connected layers with ReLU as activation function with one neuron in output layer (b) Actor Network: two hidden layers with ReLU activation, followed by mean and sigma estimation layer at the same level. These estimations are used for creating distributions from which actions are sampled (c) Hyper-parameters for actor and critic networks}
\end{figure*}

\subsection{Critic Network}\label{critic_arc}
Critic network(Q-function) is used to estimate the value of a state, that is,  the expected reward from this state. Critic network consists of three hidden layers and an output layer with one neuron. Fig \ref{fig:critic_neteork} shows the architecture of the network. Hyper parameters for the critic network is shown in Figure \ref{fig:hyperparam}.

\subsection{Actor Network}\label{actor_arc}
Actor network is also know as the policy network and is used to estimate the action from a given observation state. The output are the change is PWM values with respect to the current PWM values. As this is stochastic policy, we generate a normal distribution and sample the action from that distribution. We are using fully-connected layers to estimate the mean and standard deviation for each rotor. These values are used to form the normal distribution corresponding to the rotor. The sample is then passed to the hyperbolic tangent function to map the value in range of $(-1, 1)$. We rescale this value to change in PWM signal. As the change in hardware control signals cannot be very high, we have limited the maximum change of PWM to 0.15 at any step. We have a linear mapping from $(-1, 1)\rightarrow(-0.15, 0.15)$.  A depiction of the network is presented in Fig. \ref{fig:actor_network}. Hyper-parameters for the actor network is shown in Figure \ref{fig:hyperparam}.
\subsection{Training}
The algorithm takes two critic networks or Q-functions, $\theta_1$, $\theta_2$  and one actor network or policy function $\phi$. The parameters of each Q-function is copied to the target function. Replay buffer $D$ is initialized as empty. Now, for each step, we start by sampling an action $a$ for the current state $s$. We pass the $a$ to simulator and observe the next state $s'$, reward $r$ and if the episode has ended, i.e if $d$ is true. We store the transition $(s, a, r, s', d)$ in the $D$. As this algorithm is off-policy, to update the networks, we sample a batch $B$ of transitions from $D$.  The loss function $\mathbb{L}(\theta_i, B)$ of Q-network $Q_i$~with~parameters~$\theta_i$ in SAC is given as
\begin{equation}\label{eqn:Q-Loss}
    \mathbb{L}(\theta_i, B) = \mathbb{E}_{(s, a, r, s', d)\in B} \left[(Q_{\theta_i}(s, a) - \hat{Q}(r, s', d))^2\right],
\end{equation}
where the target $\hat{Q}(r, s', d)$ is given by 
\begin{equation}\label{eqn:target_calc}
    \hat{Q}(r, s', d) = r + \gamma(1-d)(\min_{i=1,2} Q_{\bar{\theta}_i} (s', a) - \alpha\log\pi_\phi(a'| s')),
\end{equation}
where~a' $\sim \pi_\phi$.
We update both the Q-functions by one step of gradient descent as given by 
\begin{equation}\label{eqn:Q-update}
    \nabla_{\theta_i} \frac{1}{|B|} \sum_{(s, a, r, s', d)\in B} (Q_{\theta_i}(s, a) - \hat{Q}(r, s', d))^2.
\end{equation}

Similarly, to learn the policy network, we need to maximize the expected reward and entropy. The expectation of policy can be written as 
\begin{equation}\label{eqn:policy_expectation}
    \mathbb{E}\left[(Q_{\bar{\theta}_i} (s, a) - \alpha\log\pi_\phi(a| s))\right].
\end{equation}
As we want to maximize the reward, gradient ascent is determined as 
\begin{equation}\label{eqn:policy_update}
    \nabla_{\phi}\frac{1}{|B|}(\min_{i=1,2} Q_{\bar{\theta}_i} (s, a_\phi (s)) - \alpha\log\pi_\phi(a_\phi (s)| s))
\end{equation}
where $a_\phi (s)$ is sampled from the new policy. We can also perform typical gradient descent update just by using the negative of equation \eqref{eqn:policy_update}. As the actions stored in $D$ are old, we sample new action $a_\phi(s)$ for each state $s$ from the current policy. These new actions are then used to update the parameters $\phi$.

Temperature coefficient $\alpha$ is one of the important hyper-parameters of SAC. $\alpha$ is updated using the methods given in \cite{SAC_temperature}
The final step of each iteration is to update the target networks. The target Q-networks are obtained by polyak averaging the Q-network parameters in an iterative manner. $\rho$ is step size used for polyak averaging. The complete flow of SAC is shown in Algorithm \ref{alg:sac}.

\begin{algorithm}
\SetAlgoLined
$INPUT: \theta_1, \theta_2, \phi$ \\
$\bar{\theta}_1 \gets \theta_1, \bar{\theta}_2 \gets \theta_2$ \\
$D \gets \emptyset$ \\

\For{each environment step}{
    $a \sim \pi_\phi(.|s)$ \\
    $s' \sim p(s'|s, a)$ \\
    $D \gets D \bigcup \{(s, a, r, s', d)\} $ \\
    $Sample~a~batch,~B={(s, a, r, s', d)}~from~D$ \\
    $\theta_i \longleftarrow \theta_i - \lambda_Q \hat{\nabla}_{\theta_i} J_Q(\theta_i)~~for~i \in \{1, 2\}$ \\
    $\phi \gets \phi - \lambda_\pi \hat{\nabla}_\phi J_\pi(\phi)$ \\
    $\alpha \gets \alpha - \lambda \hat{\nabla}_\alpha J(\alpha)$ \\
    $\hat{\theta_i} \gets \rho\theta_i - (1-\rho)\hat{\theta}_i ~~for~i \in \{1, 2\}$ }
$OUTPUT:\theta_1, \theta_2, \phi$\\
\caption{Soft Actor-Critic Algorithm}
\label{alg:sac}
\end{algorithm}



\begin{figure*}
\centering
\subfloat[\label{Reward_plot}]{\includegraphics[width = 6cm,height=4cm]{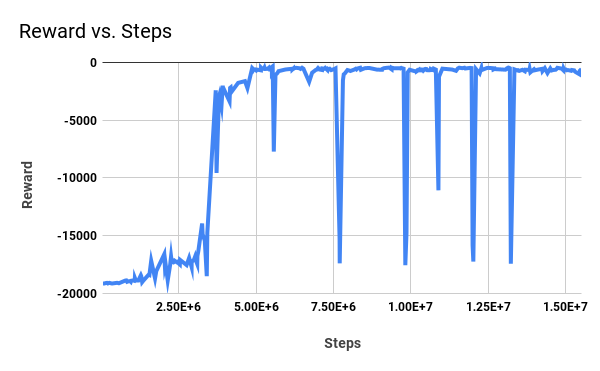}}
\subfloat[]{\includegraphics[width =6cm,height=4cm]{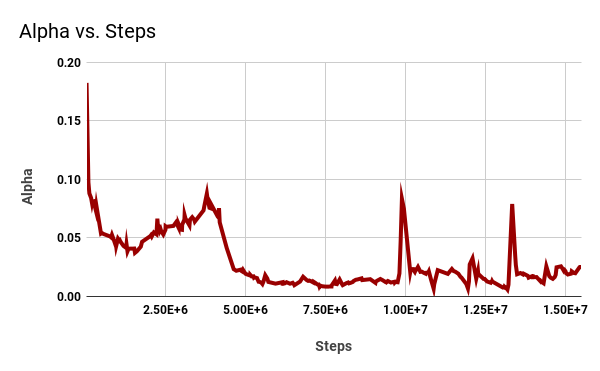}}
\subfloat[]{\includegraphics[width = 6cm,height=4cm]{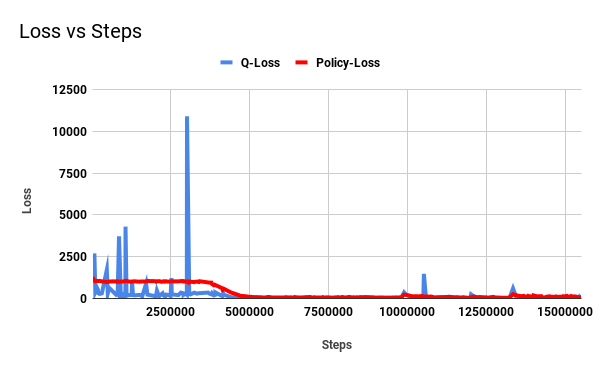}}
\caption{a) Rewards vs Steps plot, Policy network achieved high reward value in 5 Million steps or 5000 episodes. We let the policy train to converge further after that. b) $\alpha$ vs Steps plot, It can be seen that $\alpha$ starts with a high value which means high exploration, The $\alpha$ converges as rewards converges which leads to low exploration and high exploitation of the learned policy. c) Loss vs Steps. Q-Loss and Policy loss convergence is in sync with the convergence of rewards and $\alpha$. Each of the plots are compiled over 15 Million steps}
\vspace{-2pt}
\label{Training plots}
\end{figure*}

\section{Simulation results} \label{sec:results}

\subsection{Simulator Setup}
The proposed algorithm is evaluated using a python based quadrotor simulator \cite{bobzwikQ47:online}. 
 The simulator design allows us to pass the PWM values to rotors, which is then mapped to the rotor's RPM by the simulator, and simulate it for $\Delta t$ time. All the values that's required for the observation state is taken from the simulator and passed to the SAC algorithm to process. An action is generated by the actor network which is then simulated to achieve the next state of the quadrotor. We used a control frequency of 100Hz, i.e. $\Delta t = 0.01sec$ for running the simulator. 
 All the parameters of the quadrotor used in the simulator is shown in Table \ref{tab:sim_param}.
\begin{table}
    \centering
    \begin{tabular}{|c|c|}
    \hline
        Mass of quadrotor ($kg$) & 1.2 \\
        Max RPM & 900 \\
        Min RPM & 0 \\
        Max thrust per rotor ($N$) & 9.1 \\
        Arm length ($m$) & 0.16 \\
        Motor height ($m$) & 0.05 \\
        Rotor moment of inertia ($kg*m^2$) & 2.7e-5 \\
        Inertial tensor of the quadrotor ($kg*m^2$) & $\begin{bmatrix}
            0.0123 & 0 & 0 \\
            0 & 0.0123 & 0 \\
            0 & 0 & 0.0123
        \end{bmatrix}$ \\
        Thrust coefficient ($N/(rad/s)^2$) & $1.076 \times 10^{-5}$\\
        Torque coefficient ($Nm/(rad/s)^2$) & $1.632 \times 10^{-7}$\\
        
    \hline
    \end{tabular}
    \caption{Parameters used in the simulator while training}
        \vspace{-20pt}
    \label{tab:sim_param}
\end{table}

\subsection{Training and Testing}

\begin{figure*}
\centering
\subfloat[Hovering: Coordinates]{\label{fig:hover-pose}
\includegraphics[width=6cm,height=3.5cm]{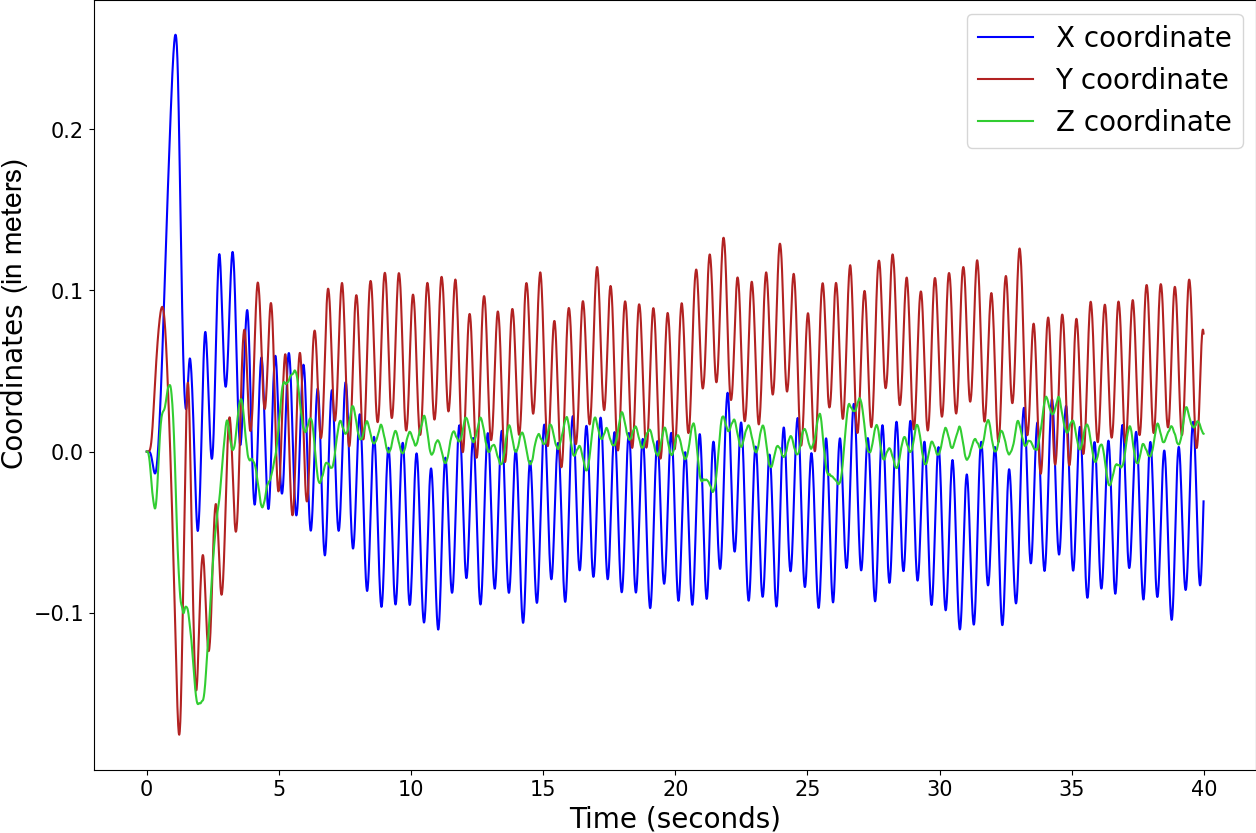}}
\subfloat[Hovering: Motor PWMs]{\label{fig:hover-thrust}
\includegraphics[width=6cm,height=3.5cm]{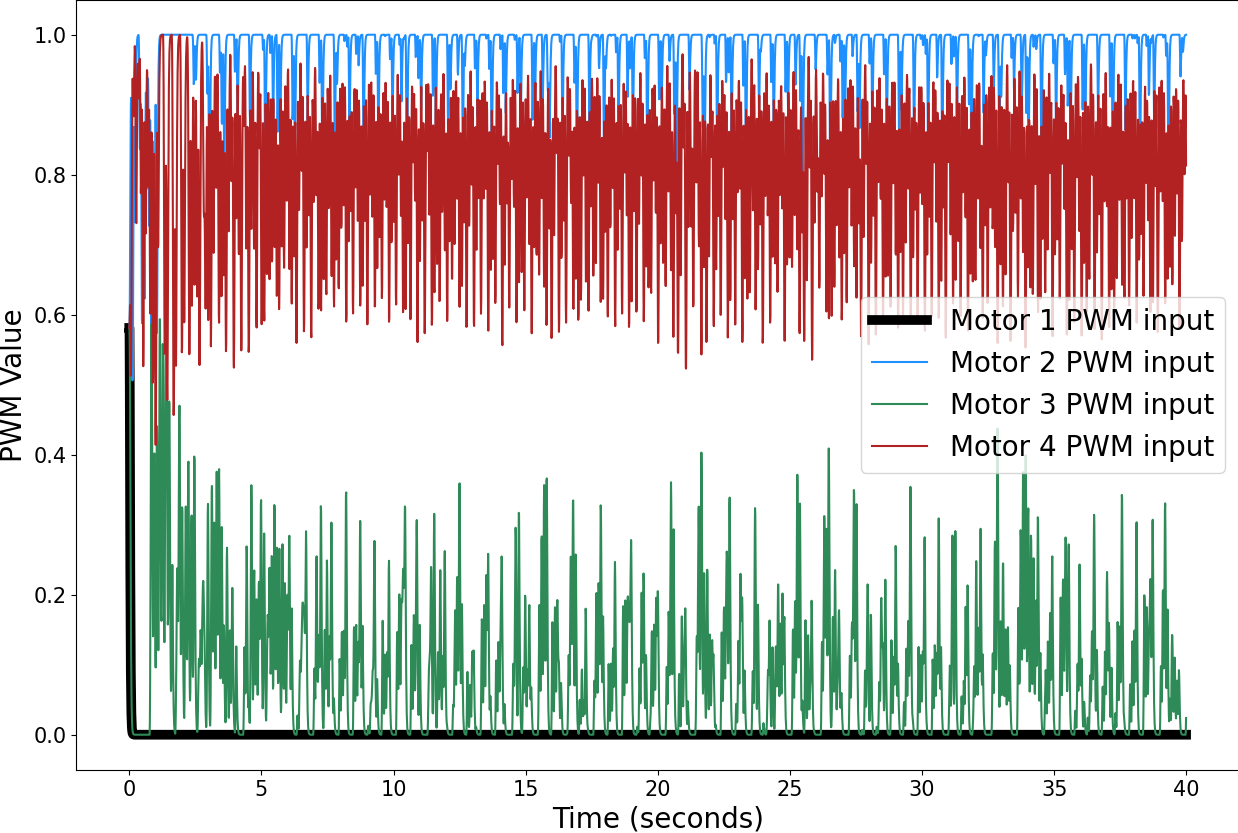}}
\subfloat[Hovering: 3D trajectory]{\label{fig:hover-sim}
\includegraphics[width=6cm,height=3.5cm]{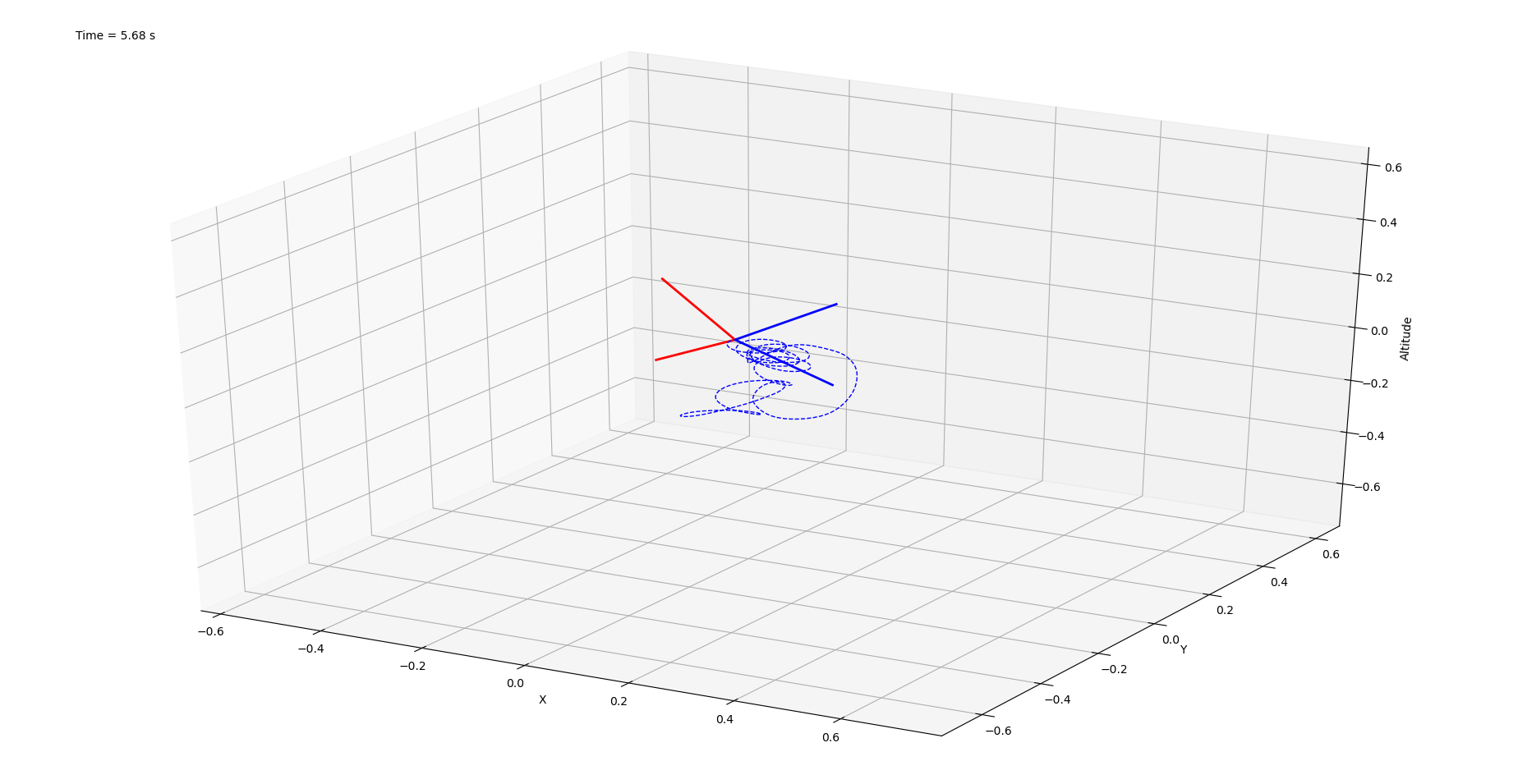}}
\newline
\subfloat[Landing: Coordinates]{\label{fig:land-pose}
\includegraphics[width=6cm,height=3.5cm]{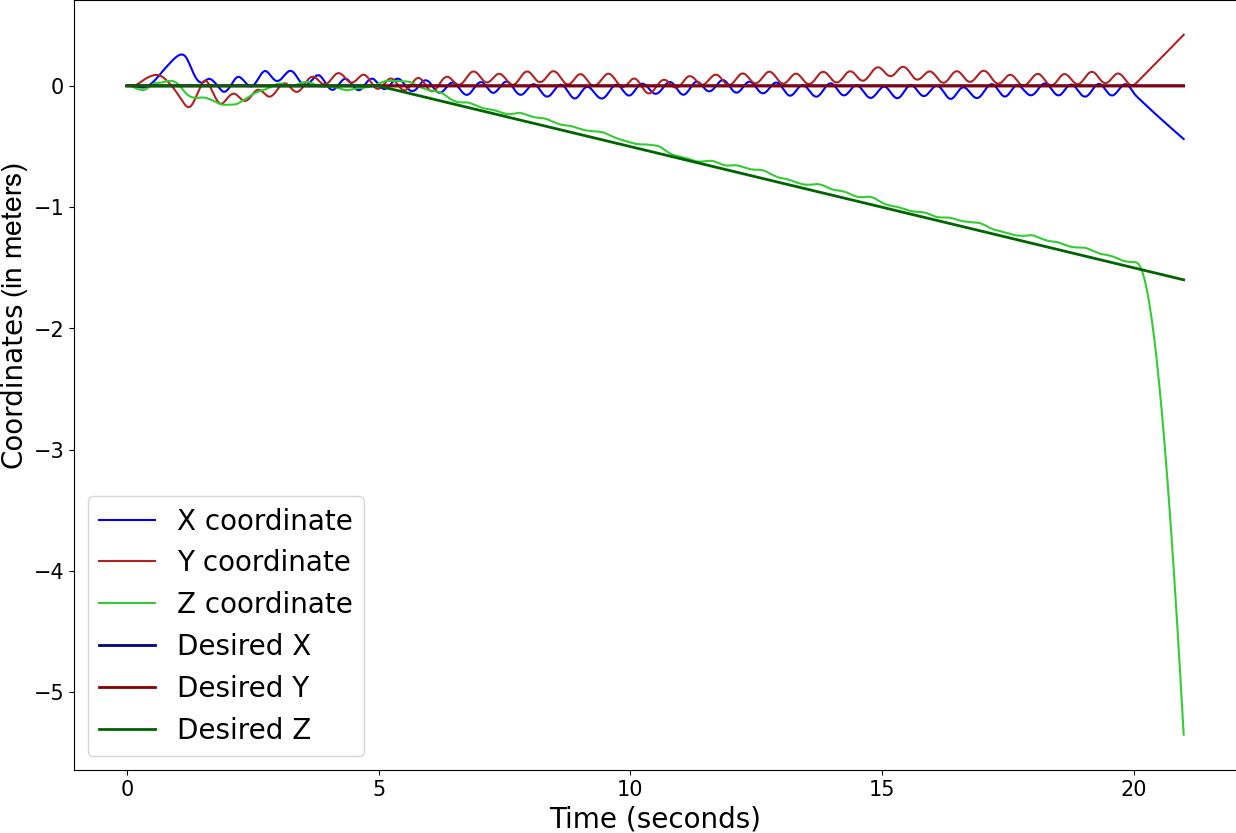}}
\subfloat[Landing: Motor PWMs]{\label{fig:land-thrust}
\includegraphics[width=6cm,height=3.5cm]{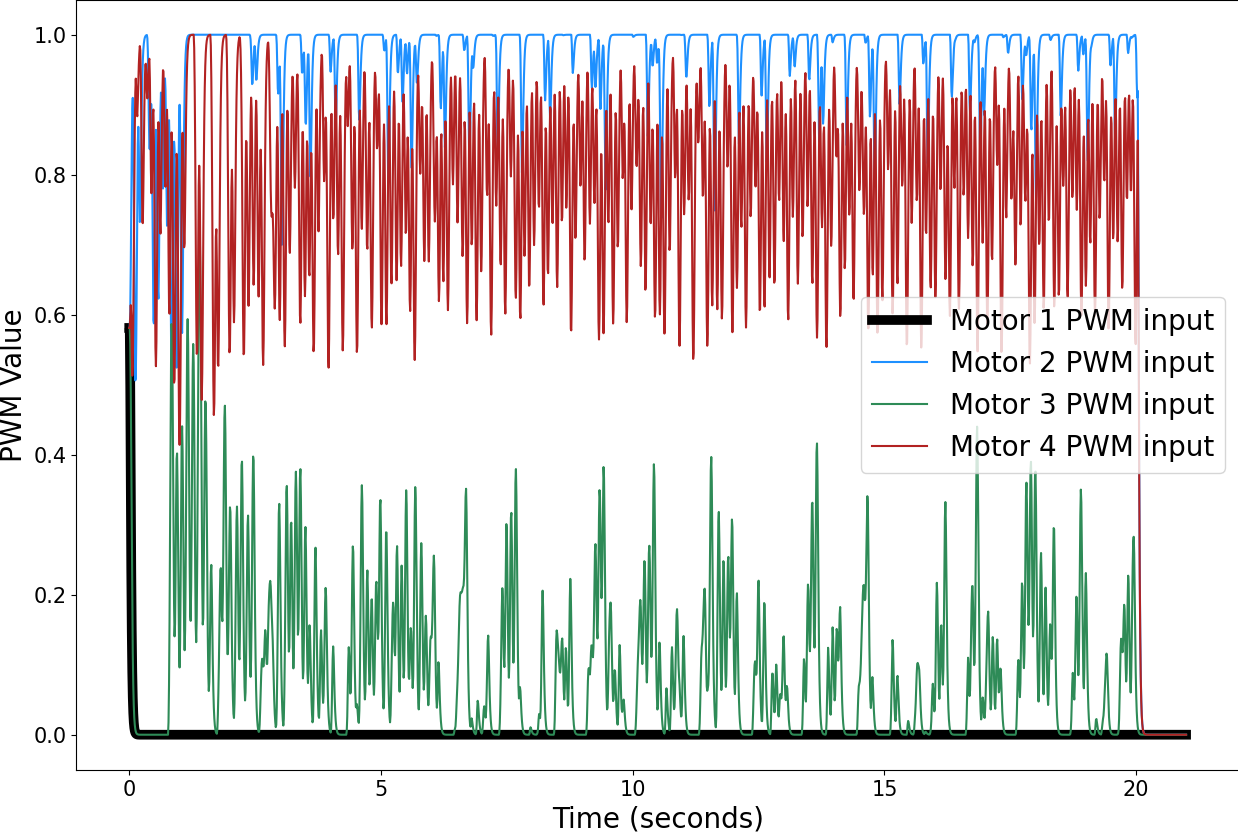}}
\subfloat[Landing: 3D trajectory]{\label{fig:land-sim}
\includegraphics[width=6cm,height=3.5cm]{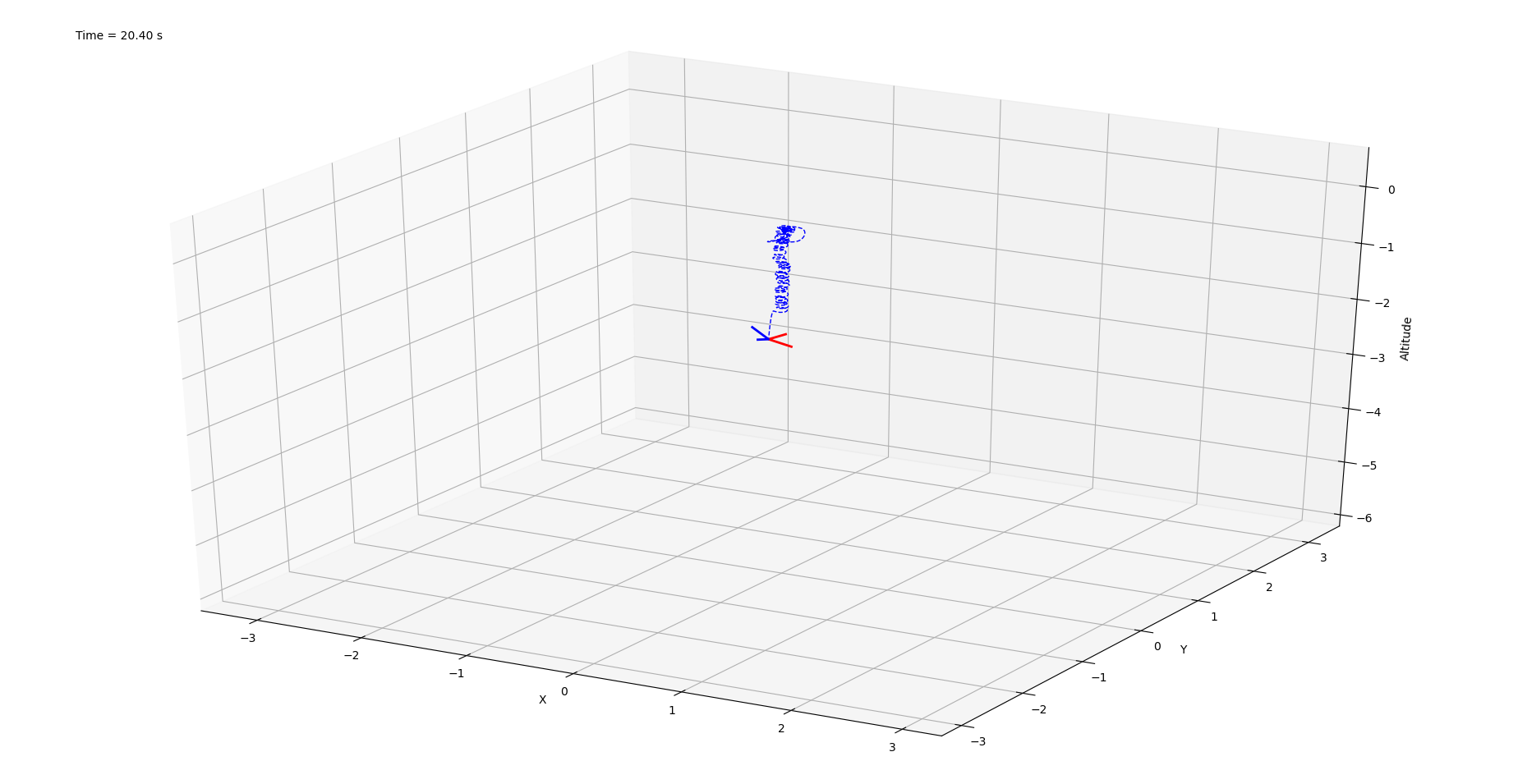}}
\newline
\subfloat[XY circle: Coordinates]{\label{fig:xy-pose}
\includegraphics[width=6cm,height=3.5cm]{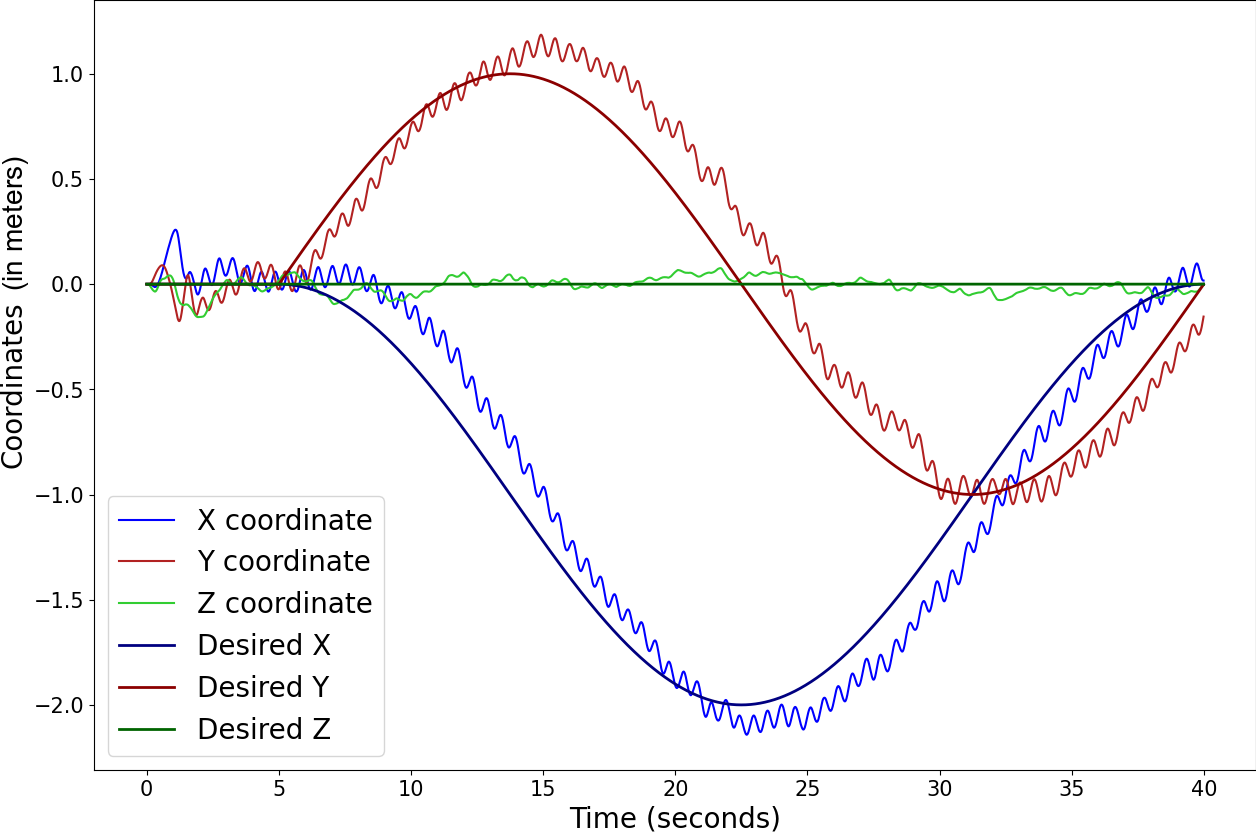}}
\subfloat[XY circle: Motor PWMs]{\label{fig:xy-thrust}
\includegraphics[width=6cm,height=3.5cm]{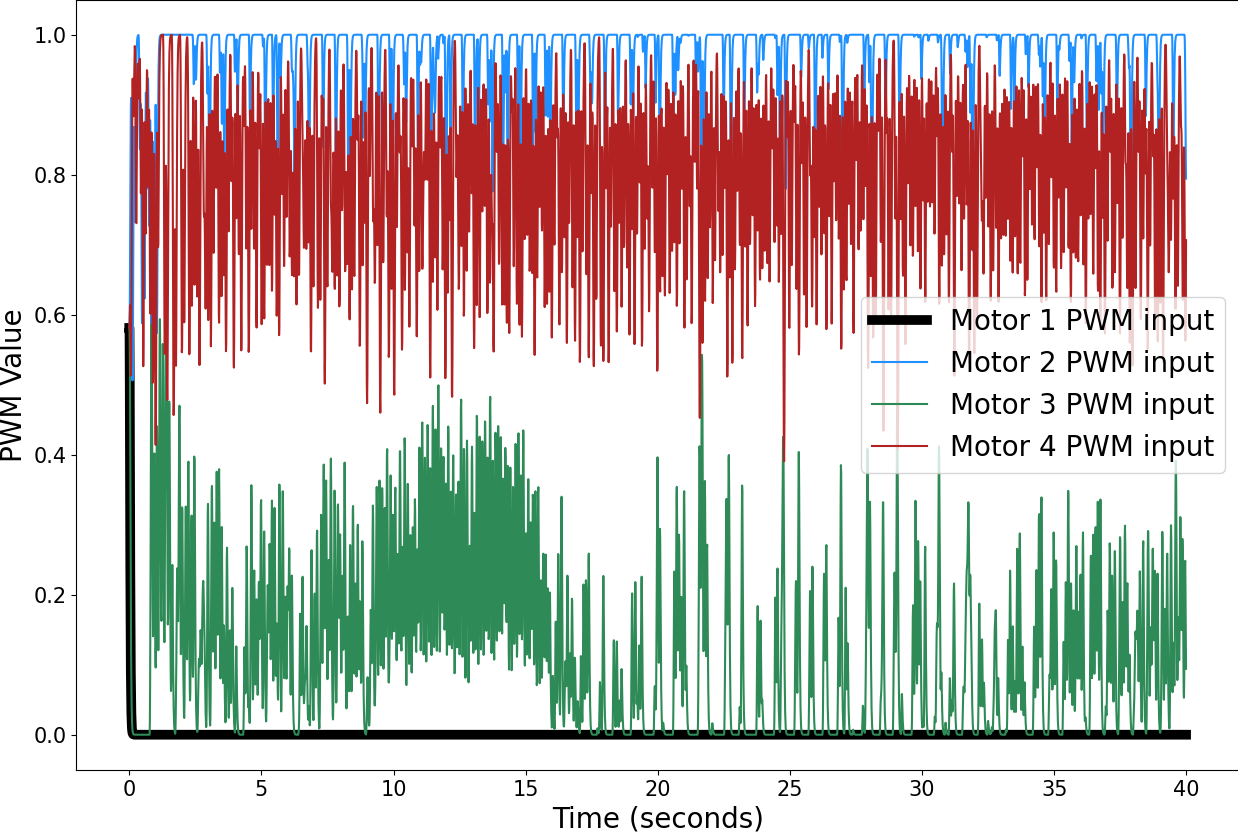}}
\subfloat[XY circle: 3D trajectory]{\label{fig:xy-sim}
\includegraphics[width=6cm,height=3.5cm]{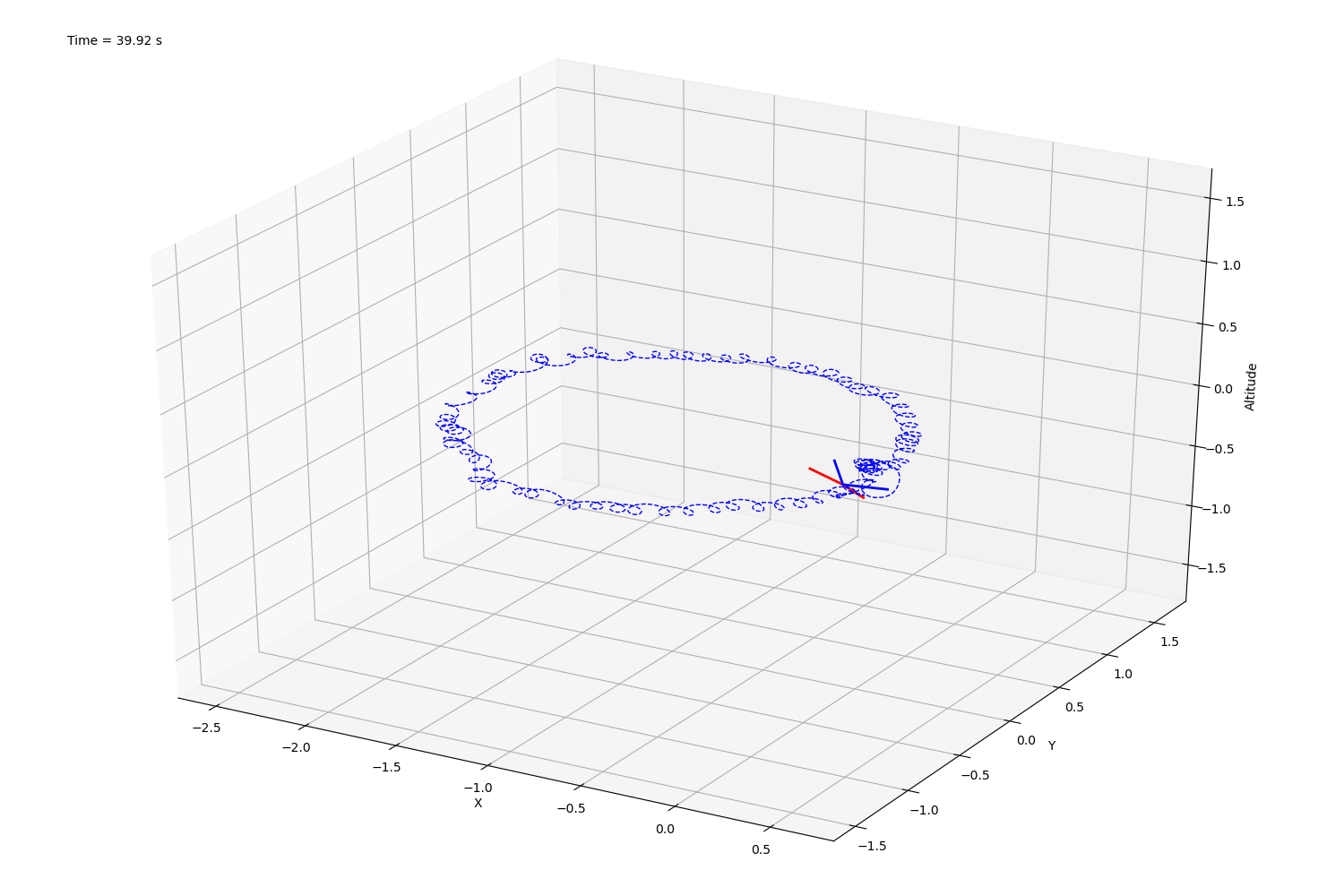}}
\newline
\subfloat[YZ circle: Coordinates]{\label{fig:yz-pose}
\includegraphics[width=6cm,height=3.5cm]{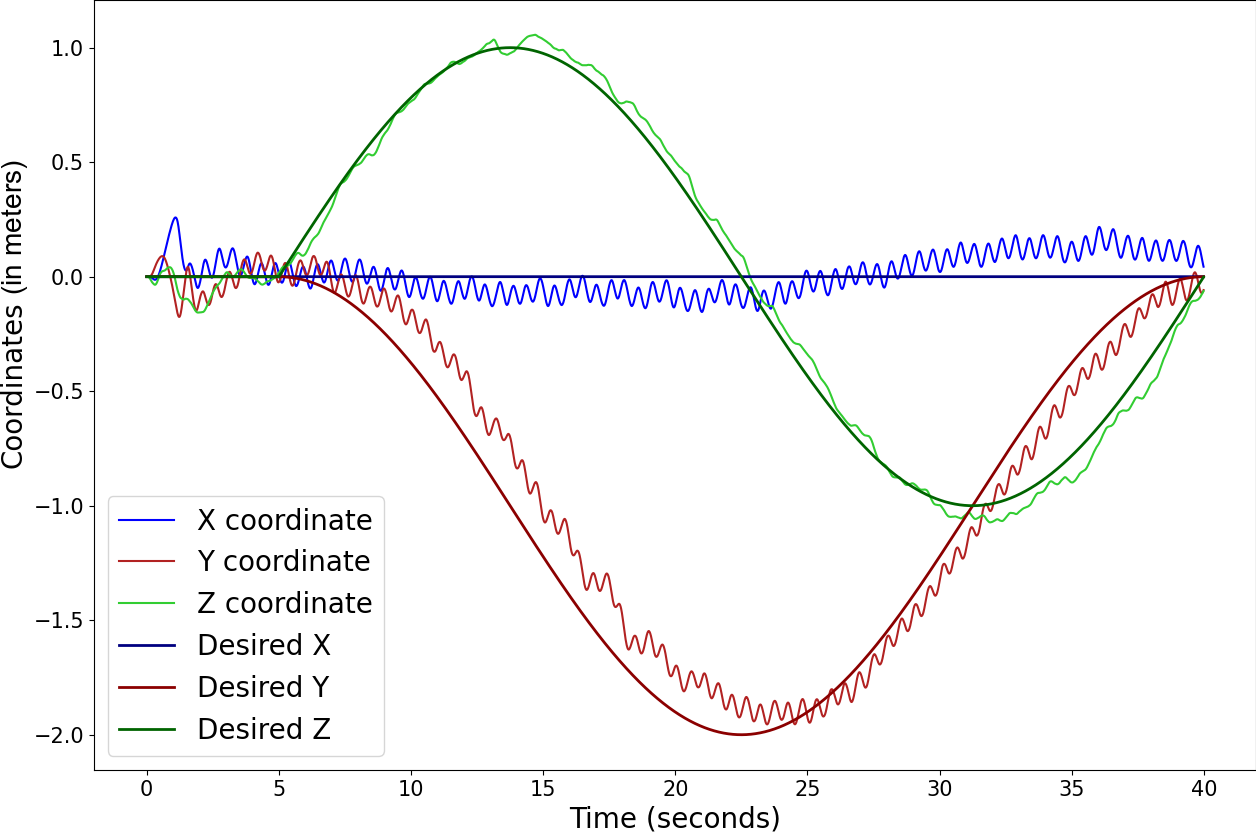}}
\subfloat[YZ circle: Motor PWMs]{\label{fig:yz-thrust}
\includegraphics[width=6cm,height=3.5cm]{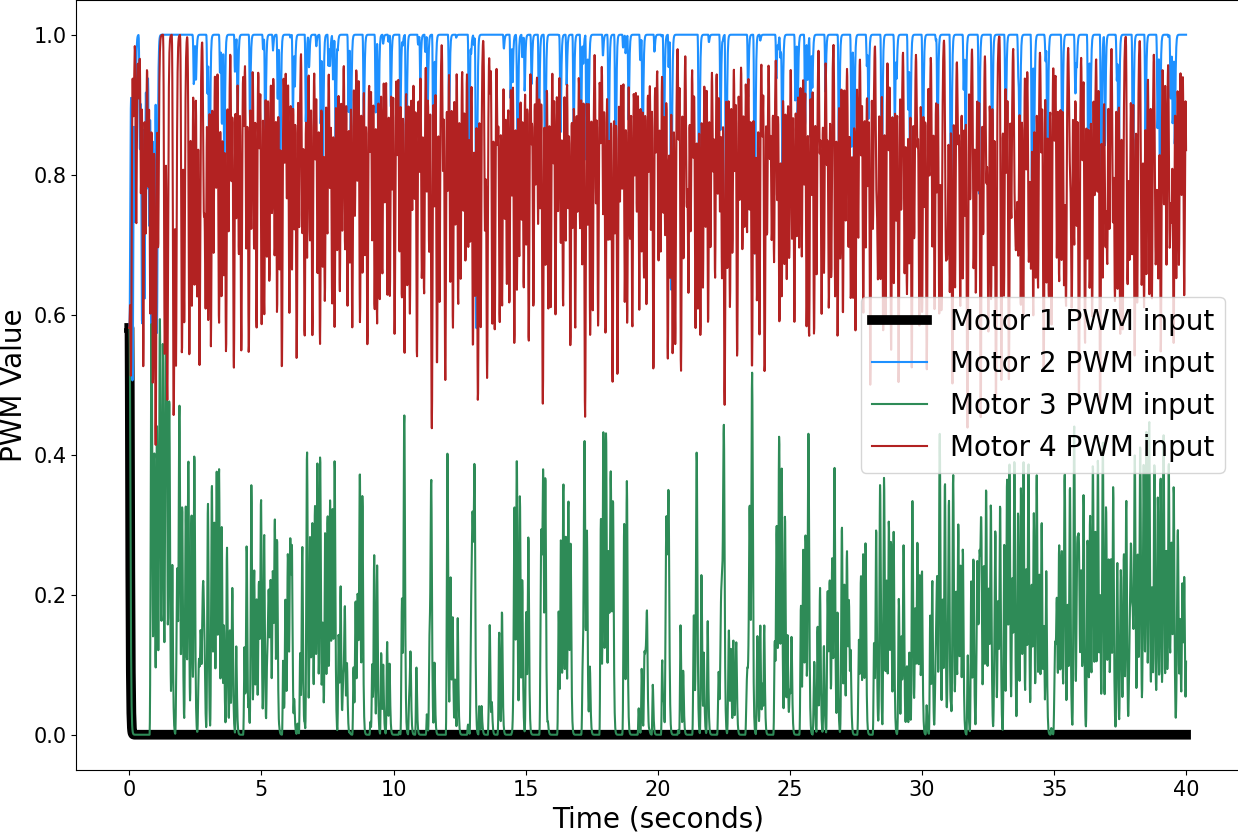}}
\subfloat[YZ circle: 3D trajectory]{\label{fig:yx-sim}
\includegraphics[width=6cm,height=3.5cm]{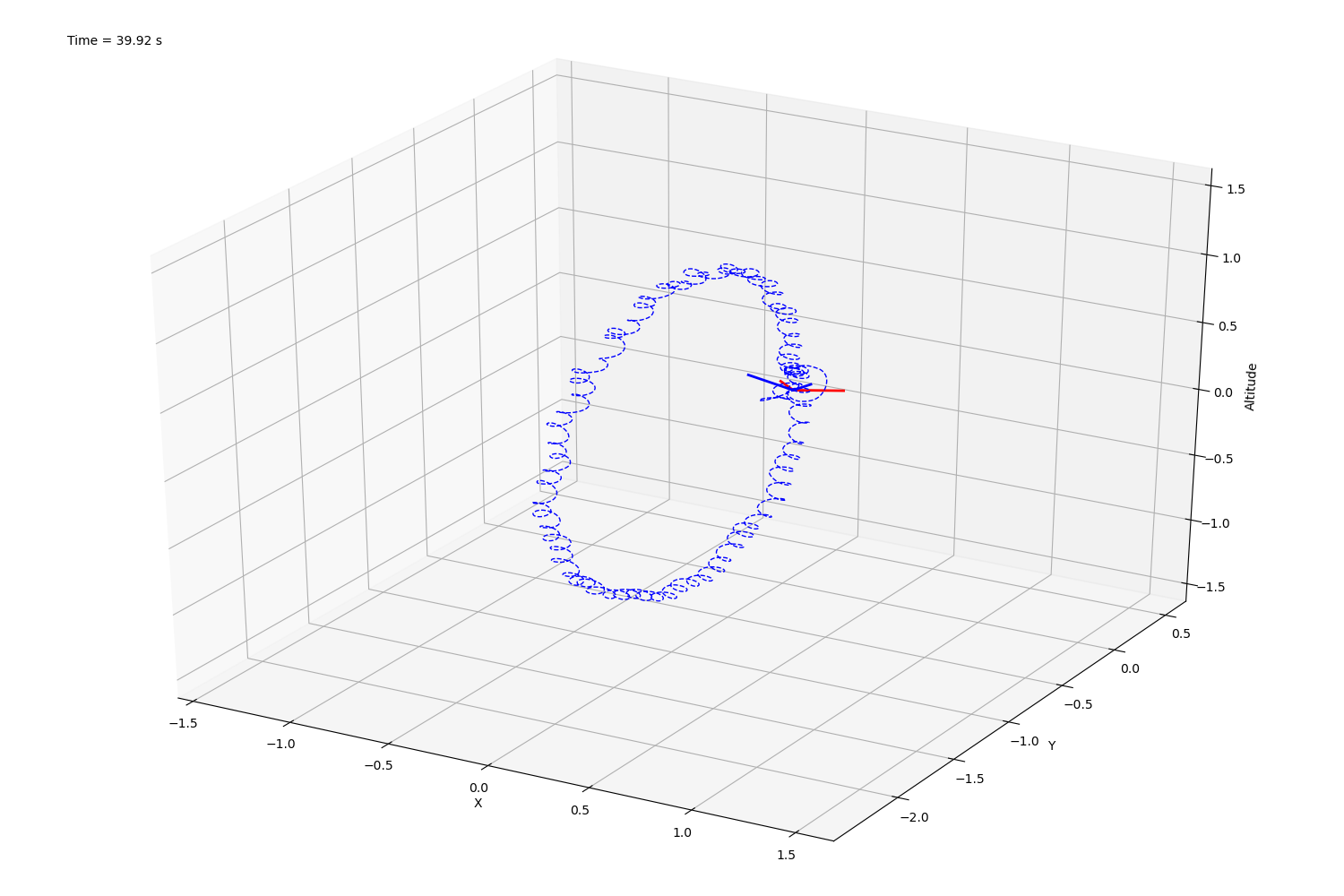}}
\newline
\subfloat[Saddle: Coordinates]{\label{fig:xyz-pose}
\includegraphics[width=6cm,height=3.5cm]{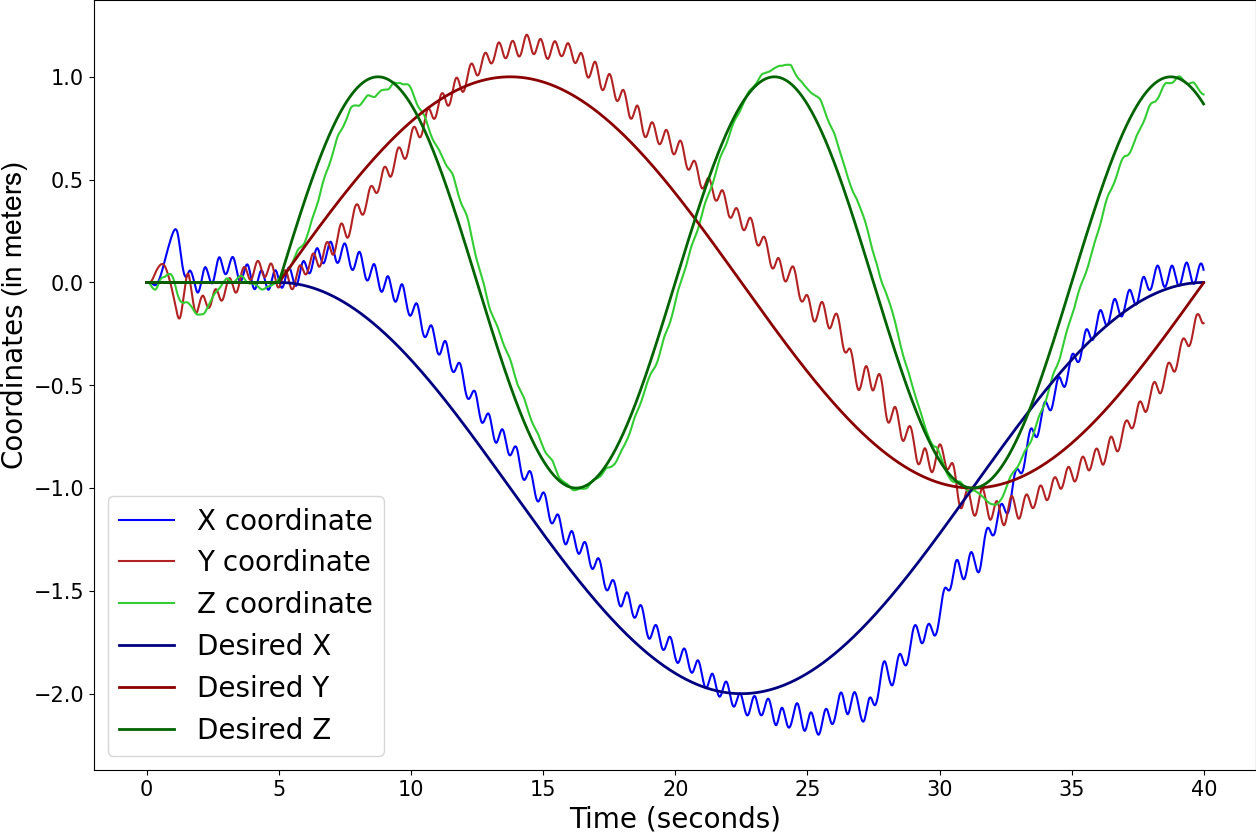}}
\subfloat[Saddle: Motor PWMs]{\label{fig:xyz-thrust}
\includegraphics[width=6cm,height=3.5cm]{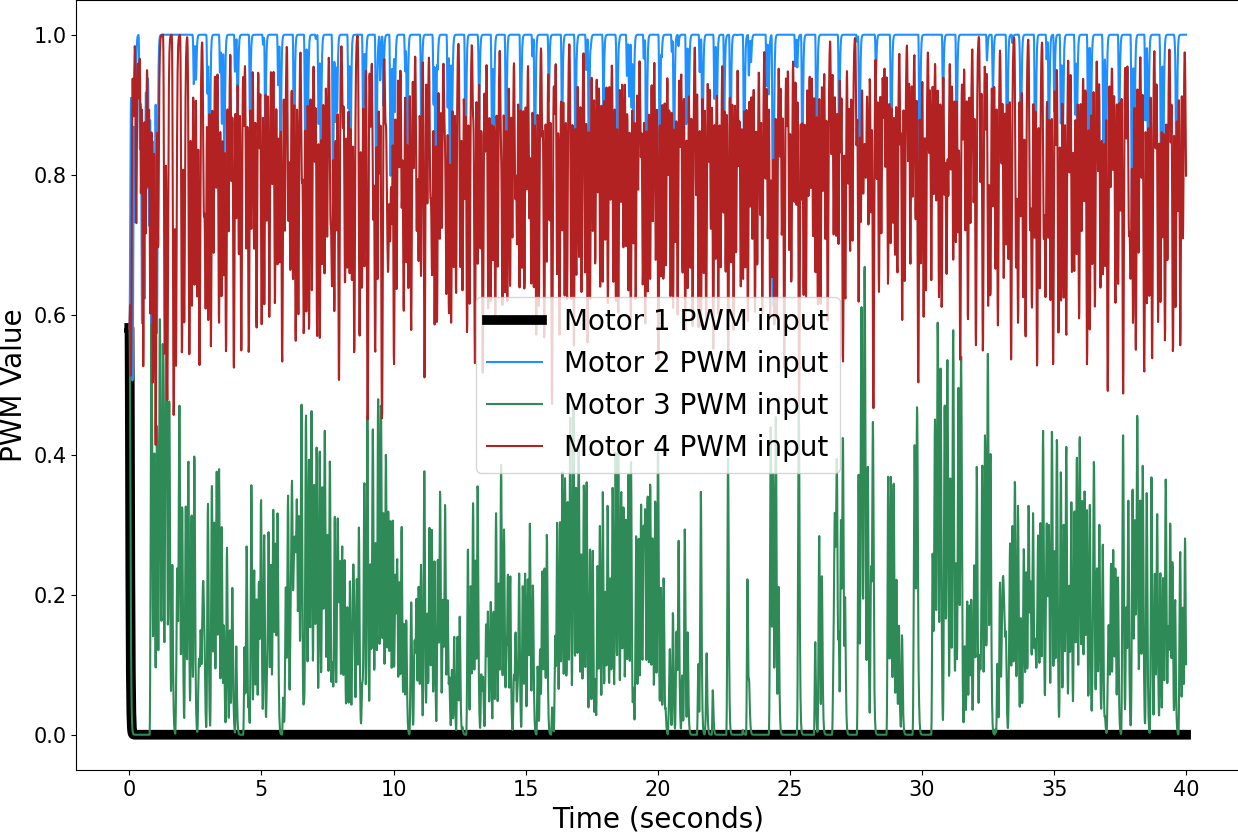}}
\subfloat[Saddle: 3D trajectory]{\label{fig:xyz-sim}
\includegraphics[width=6cm,height=3.5cm]{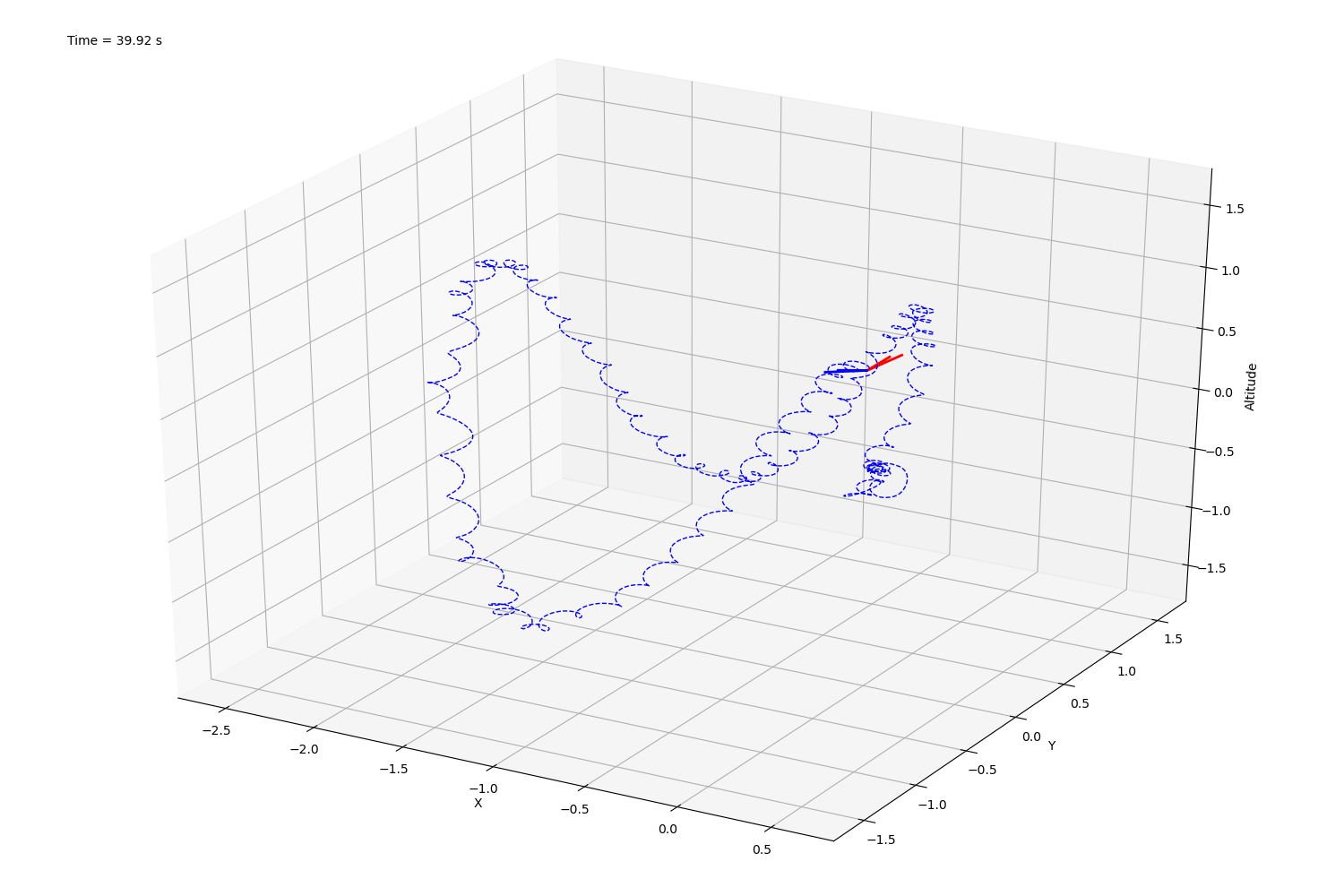}}
\caption{Test run results. $1^{st}$ row: hovering maneuver, $2^{nd}$ row: landing maneuver, $3^{rd}$ row: circular trajectory in XY plane, $4^{th}$ row: circular trajectory in YZ plane, $5^{th}$ row: saddle shaped trajectory. 'Coordinates' plots the actual XYZ coordinates of the quadrotor as well as the desired XYZ coordinates to perform the maneuver. 'Motor PWMs' plots the PWM input to every rotor. One of the rotos never spins and its angular velocity can be seen to be 0 in all the plots. '3D trajectory' shows the the actual trajectory maneuvered by the quadrotor in 3D space.} \label{fig:test-results}
\end{figure*}

We used 10 seconds of simulation for each episode of training with a control frequency of 100Hz, resulting in every episode having 1000 steps. Every episode of training and testing start with the quadrotor hovering at the origin with all rotors working. We then disable one of the rotors and let  the SAC DRL algorithm generate the PWM values for the remaining 3 rotors. The goal position is set to be the origin itself. 

The hyper-parameters of the algorithm are $\gamma=0.99,\rho=0.05,\lambda_{\theta_i}=1,\lambda_{\phi}=1,\lambda=1$ and Replay Buffer size of 1e6 and Batch size is of 256. 
We trained the algorithm for about 15 Million steps with the actor and critic being updated at each step. The various plots achieved during training are shown in Fig \ref{Training plots}. In Fig \ref{Reward_plot}, we can see sudden spikes in the rewards, which are a consequence of exploration of new states by the quadrotor. In the algorithm, given a state, we take a random action with probability of 0.001 which is a part of exploration strategy. The quadrotor starts at a fixed position, so, the space explored will be limited. As the policy tends to converge, probability of exploring new states also reduces. But exploring new state can lead to few spikes in the rewards.

Test simulations were run for 40 seconds with control frequency of 100Hz. We tested 3 types of maneuvers on a quadrotor with a single rotor failure: \textit{(a)} hovering, \textit{(b)} landing and \textit{(c)} path following in 3D space.

\subsection{Hovering}

In this test the vehicle is initialized at the origin and made to hover there by setting the goal as the origin itself. In the real world, the origin shall be shifted to the altitude at which the quadrotor should hover. In this and the subsequent tests, \textit{rotor 1} of the vehicle have been disabled in order to mimic a SRF. Fig.\ref{fig:hover-sim} shows the trajectory of the quadrotor in 3D space, while hovering at the origin. Slight swinging behaviour can be seen in the position of the quadrotor as it tries to compensate for the failed rotor. Fig.\ref{fig:hover-pose} shows the XYZ coordinates of the quadrotor vs. the time elapsed. Periodic oscillations along with slight deviation from the origin can be seen in the X and the Y coordinates caused due to the loss of yaw control which leads the quadrotor to rotate about it's vertical axis. The PWM inputs shown in Fig.\ref{fig:hover-thrust} indicates that the model learnt to slow down the \textit{rotor 3} that is diagonally opposite to \textit{rotor 1}, to keep the net torque on the UAV close to 0. Ideally, the diagonally opposite rotor should also be set to 0 while the other 2 rotors should rotate with the same RPM but that is only feasible when the vehicle is already in a perfectly horizontal position at the time of occurrence of the fault and it does not need to make any roll and pitch corrections (and doing this would also mean that the quadrotor would lose maneuvering capabilities). Our algorithm on the other hand, learns to use all the three remaining rotors to be able to correct for orientation errors as well to retain its maneuverability. It uses  \textit{rotor 3} intermittently to provide necessary torques for roll and pitch controls. It also oscillates the speed of \textit{rotor 2} and \textit{rotor 4} to maintain orientation and roll and pitch control. \textit{Rotor 2} and \textit{4} also increase their speed up to be able to support the weight of the quadrotor. 

\subsection{Landing}

In this test, the quadrotor is initiated at the origin with a failure of \textit{rotor 1}. The goal is then set at the origin itself for the initial 5 seconds, in order to let the quadrotor recover from the failure and stabilize the hovering motion at its current position. Then the altitude of the goal is slowly reduce at the rate of $10 cm/s$, leading the quadrotor to start its descent. Since the simulator we use, lacks a physical ground, we assume that the ground is 1.5 meters below the initialization position (or the origin). Upon descending for 1.5 meters, the inputs to all the rotors are cut-off, resembling the standard procedure of landing a quadrotor where the thrust is cut off just before touching down on the ground. Fig.\ref{fig:land-pose} shows that the quadrotor performs a controlled descent by maintaining the descent rate until all the rotors are shut down. In this case, \textit{rotor 3} is used less frequently than for any other maneuver (Fig.\ref{fig:land-thrust}). The descent trajectory followed by the quadrotor is shown in Fig.\ref{fig:land-sim}.

\subsection{Path Following}

This test shows that our algorithm can provide maneuverability to a quadrotor with a single rotor failure. We performed three tests to show the robustness of our algorithm against any trajectory: \textit{(a)} moving in a circular trajectory in the XY plane, \textit{(b)} moving in a circular trajectory in the YZ plane and \textit{(c)} moving in a saddle shaped trajectory spanning the entire 3D space.
The results for the same can be seen in  Fig.\ref{fig:xy-pose}-\ref{fig:xy-sim}, Fig.\ref{fig:xyz-pose}-\ref{fig:xyz-sim} and Fig.\ref{fig:yz-pose}-\ref{fig:xyz-sim}. For all the 3 trajectories, the quadrotor is initialised at the origin and the for the first 5 seconds, the goal position is set to be the origin itself. This is done to give the quadrotor some time to recover from the fault and stabilize itself. The the goal position is slowly moved along the desired trajectory, making the quadrotor follow the trajectory. From the 'Coordinates' graphs of Fig.\ref{fig:test-results} we can see that the quadrotor successfully follows the desired trajectory, with only a slight deviation from the desired path.

\subsection{Wind Tolerance}

\begin{figure}
\centering
\subfloat[Saddle: Coordinates]{\includegraphics[clip,trim=0mm 0mm 5mm 0mm, width = 4.25cm]{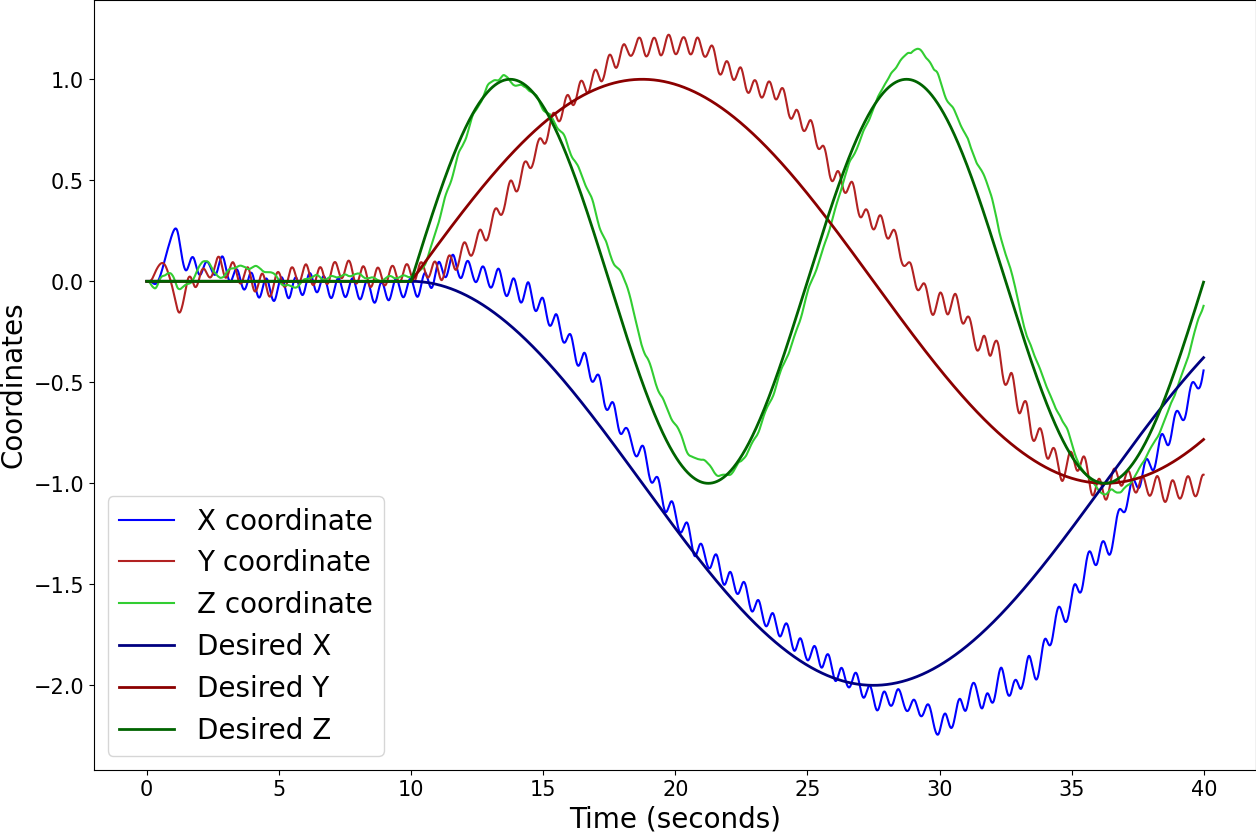}}\label{pwm_wind} 
\subfloat[Saddle: Simulation]{\includegraphics[clip,trim=46mm 5mm 5mm 5mm, width = 4.25cm]{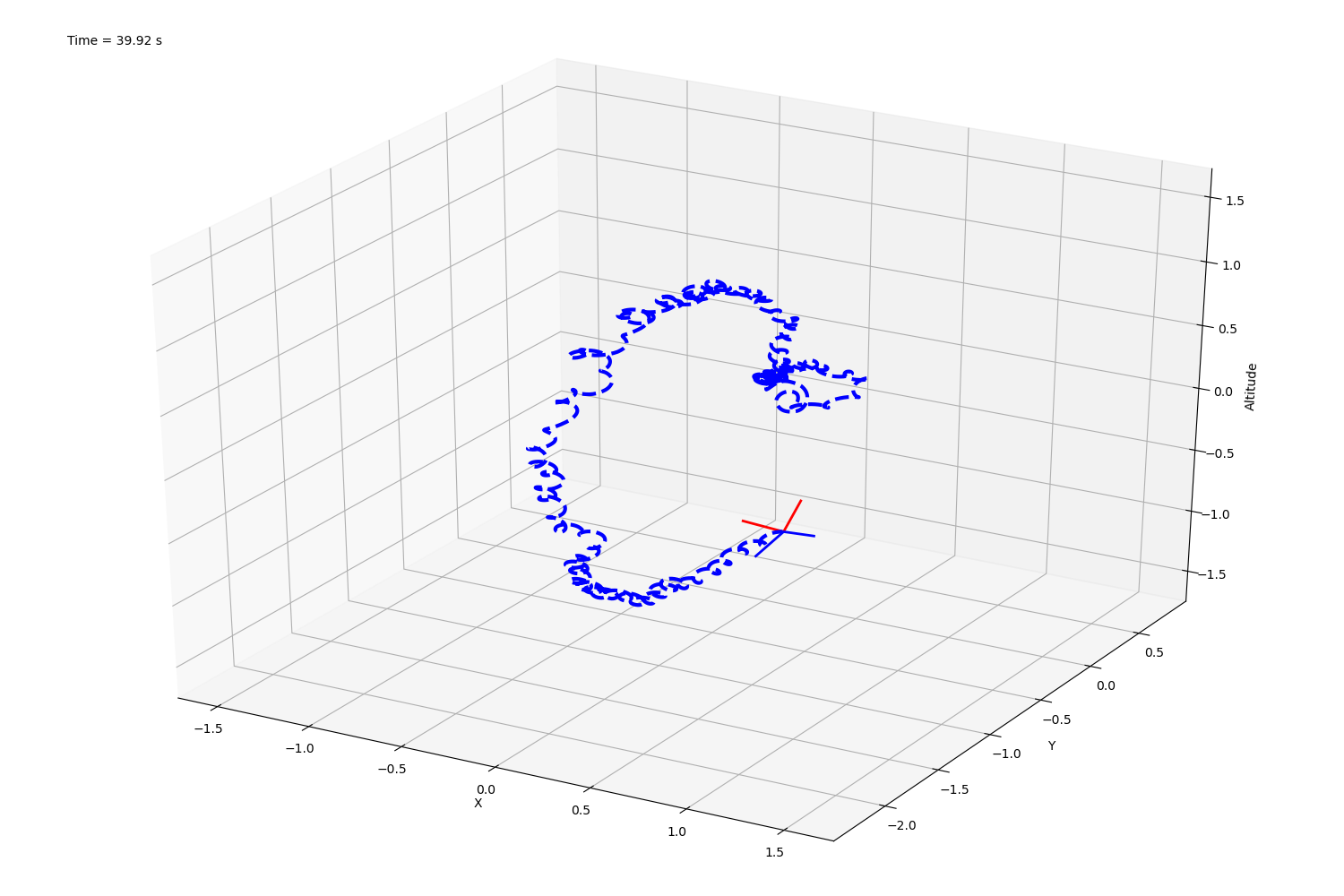}}\label{pos_wind} 
\caption{Results with added wind disturbance}
\label{Wind model sim}
\end{figure}

To show the tolerance against wind, we added a wind model with random directions. We found that the quadrotor was able to tolerate winds with speed up to $2 m/s$. This shows that the model is robust against light breezes even though while training, wind disturbances were not included. In the presence of winds, we need to provide more time for the quadrotor to first stabilize from the SRF than without wind This can be seen in Fig.\ref{Wind model sim}. We have presented the results for motion of quadrotor following the saddle shaped trajectory under these wind conditions. Here it can be seen that drone needs the initial 10 seconds to stabilize its hover maneuver The quadrotor has no observation variable corresponding to the wind while training. The results show that the quadrotor is able to maneuver the trajectory in every axes robustly. 


\section{Conclusions} \label{sec:conclusions}

In this paper, we modeled and evaluated a model free deep reinforcement learning based algorithm that uses Soft Actor-Critic methods to handle a single rotor failure in quadrotors. The algorithm can stabilize a quadrotor from SRF and also provides maneuvering capabilities to the quadrotor with just 3 active rotors. The SAC-based controller was able to hover, land and following 3D trajectories. The SAC-based algorithm can run at 100Hz which is as good as a micro-controller based controller. The controller needs to be tested on a quadrotor for its actual performance assessment. The future work can be extended to study multi-robot failure, and integrate a high-level planner to maneuver to safety location for landing.

\bibliographystyle{IEEEtran}
\bibliography{references}

\end{document}